# EmbeddedML: A New Optimized and Fast Machine Learning Library

H. H. Çalışkan[1*], T. Koruk[1]

[1]Department of Computer Engineering, Faculty of Engineering and Natural Sciences, Bursa Technical University, Bursa, Türkiye

*Corresponding author: caliskanhalil815@gmail.com

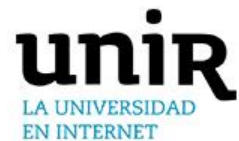

## Abstract

Machine learning models and libraries can train datasets of different sizes and perform prediction and classification operations, but machine learning models and libraries cause slow and long training times on large datasets. This article introduces EmbeddedML, a training-time-optimized and mathematically enhanced machine learning library. The speed was increased by approximately times compared to scikit-learn without any loss in terms of accuracy in regression models such as Multiple Linear Regression. Logistic Regression and Support Vector Machines (SVM) algorithms have been mathematically rewritten to reduce training time and increase accuracy in classification models. With the applied mathematical improvements, training time has been reduced by approximately 2 times for SVM on small datasets and by around 800 times on large datasets, and by approximately 4 times for Logistic Regression, compared to the scikit-learn implementation. In summary, the EmbeddedML library offers regression, classification, clustering, and dimensionality reduction algorithms that are mathematically rewritten and optimized to reduce training time.

## I. Introduction

STATISTICAL methods and machine learning algorithms enable the extraction of meaningful patterns from datasets of varying sizes and complexity. Using trained mathematical models depending on datasets, it becomes possible to make predictions or perform classifications on new data instances [1]. Machine learning models are impacting many sectors and disciplines today in terms of efficiency and automating processes. As a result of training the data obtained in the health sector with machine learning models, early detection and diagnosis of diseases that seriously threaten human life, such as diabetes, kidney stones, heart attack and breast cancer, can be effectively achieved [2]. In the economic sector, machine learning is used to increase the profitability and efficiency of commercial investments and marketing strategies [3]. It can also be used in schools to facilitate the education and learning of students [4]. Many libraries such as scikit-learn and TensorFlow have been produced to train machine learning models and perform various prediction and classification operations depending on the data [5]. As the size and complexity of datasets increase, training such models using CPUs becomes increasingly slow and difficult in such libraries. In order to overcome these difficulties, the EmbeddedML library was developed by using statistical methods and rewriting the algorithms by optimizing them from scratch. In the EmbeddedML library, mathematical operations such as matrix multiplication, transposing, inverting matrices, calculating covariance matrix, finding eigenvalues and eigenvectors are accelerated with NumPy [6]. The EmbeddedML library can work flawlessly with other libraries used in data science, such as Pandas, NumPy, and Matplotlib [7],[8],[9].

EmbeddedML library includes various machine learning algorithms such as Linear Regression, Multiple Linear Regression, Polynomial Regression, Logistic Regression, K-Nearest Neighbor (KNN), Support Vector Machines (SVM), Naive Bayes, K-Means Clustering and Principal Component Analysis (PCA) [10]. In the preprocessing phase, the EmbeddedML library supports various operations such as Min-Max Scaler, Standard Scaler, splitting the dataset into training and validation, and transforming the data into polynomial form [11]. To analyze the accuracy of regression models, the EmbeddedML library includes performance metrics such as R² (coefficient of determination), Mean Square Error (MSE), Mean Absolute Error (MAE) and Root Mean Square Error (RMSE). Metrics such as accuracy, precision, recall, and F1-score are used to assess the accuracy of classification models [12]. A confusion matrix is also used to analyze the results more thoroughly and specifically [13]. Additionally, in order to facilitate accessibility of the library, it can be downloaded via pip (Pip Installs Packages). Thanks to this lightweight and optimized library, machine learning model training, previously only possible on desktop computers with high hardware power, can now be performed efficiently and effectively on low-resource embedded systems—devices such as the Raspberry Pi, NVIDIA Jetson Orin Nano, and Orange Pi [14]. This increases the portability and accessibility of AI applications. Artificial intelligence solutions can be implemented in embedded systems with Python-based EmbeddedML instead of C-based machine learning libraries that are difficult to collaborate with other Python-based data science libraries such as Pandas, NumPy, and Matplotlib [15].

## II. Methods

### A. Regression Algorithms

Simple Linear Regression, Multiple Linear Regression and Polynomial Regression algorithms were mathematically rewritten using NumPy to reduce training time [16],[17].

*1. Simple Linear Regression*

The dataset consists of examples with only one independent variable, and these data were used to train the model along with the target (dependent) variable. In other words, dependent variables are predicted using independent variables. In this way, the line that gives us the most appropriate results depending on the independent data is drawn on the two-dimensional axis. To find the best-fit line, each data group in the dataset must be passed through the error function. The error function used in Simple Linear Regression is shown in Equation 1 [18].

$$\sum_1^k (y_k - (mx_k + n))^2 \tag{1}$$

To find the minimum value of this error function, partial derivatives are taken with respect to the variables of the slope (m) and the constant term (n) and the result of both partial derivatives is set equal to zero [19]. The partial derivative of the line with respect to the slope variable is shown in Equation 2, and the partial derivative of the line with respect to the constant term is shown in Equation 3.

$$\frac{\partial f(m,n)}{\partial m} = \sum_1^k -2x_k(y_k - mx_k - n) = 0 \tag{2}$$

$$\frac{\partial f(m,n)}{\partial n} = \sum_1^k -2(y_k - mx_k - n) = 0 \tag{3}$$

By equating these equations, the values of the slope (m) and constant term (n) variables are found. The formula for the slope (m) variable is shown in Equation 4, and the value of the constant term (n) variable is shown in Equation 5.

$$\frac{k\sum_1^k (x_k y_k) - \sum_1^k y_k \sum_1^k x_k}{k\sum_1^k (x_k)^2 - (\sum_1^k x_k)^2} \tag{4}$$

$$\frac{(\sum_1^k x_k y_k - m\sum_1^k x_k)}{k} \tag{5}$$

Thus, using the least squares formula, the model finds the most appropriate straight line slope and constant term using independent and dependent data. Based on the slope and constant term values found, the prediction is made based on the desired independent data. The calculation and mathematical representation of the error function in linear regression is illustrated in Fig. 1.

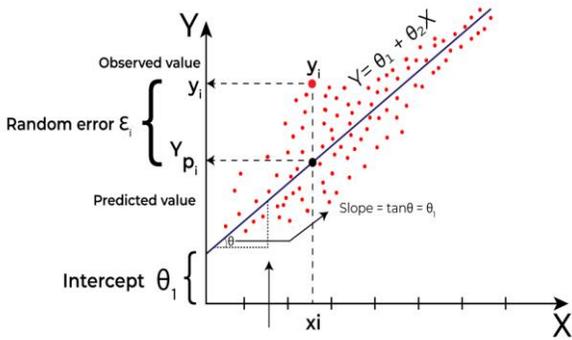

Fig. 1. Graphical representation of linear regression

Unlike the Simple Linear Regression in the Scikit-learn library, the EmbeddedML library is written in NumPy, making mathematical operations faster.

*2. Multiple Linear Regression*

In datasets with multiple independent variables, each data sample constitutes a row of the matrix. The value of 1 is added to the beginning of each row to represent the constant term. By multiplying the values of the independent variables by the coefficients (weights) corresponding to each feature, we obtain the target vector representing the dependent variables. The finding of the coefficient matrix is shown in Equation 6 [20].

$$X\beta = y \tag{6}$$

To isolate the coefficients matrix (B), we must multiply the matrix of independent data (X) by its inverse [21]. To get the inverse of the X matrix consisting of independent data, the X matrix must be a square matrix. Therefore, the X matrix is multiplied by its transpose. The result of multiplying the matrix X by its transpose is shown in Equation 7.

$$X^T X \beta = X^T y \tag{7}$$

The coefficients are found by multiplying the matrix of our independent data, which has become a square matrix, by its inverse [22]. The final form of the coefficient matrix is shown in Equation 8 and Equation 9.

$$(X^T X)^{-1} X^T X \beta = (X^T X)^{-1} X^T y \tag{8}$$

$$\beta = (X^T X)^{-1} X^T y \tag{9}$$

The dependent variable value corresponding to the new independent variable data to be predicted is calculated using the obtained coefficient matrix.

*3. Polynomial Regression*

Polynomial features are added to the matrix consisting of independent data, thus enabling the model to learn more complex relationships [23]. To add polynomial features to the independent variables, all independent variables are multiplied among themselves by taking their specific degrees and adding these new features to the original data matrix [24].

B. *Classification Algorithms*

Logistic Regression, SVM, KNN, and Naive Bayes algorithms have been mathematically modified and completely rewritten using NumPy. With these changes, the training time of the models has been shortened.

*1. Logistic Regression*

The features in each data group, consisting of a different number of independent data, are initially multiplied by weights between 0 and 1, and the bias value is added to this result. This expression is shown in Equation 10 [25].

$$z = w_1 x_1 + w_2 x_2 + w_3 x_3 + \cdots + b \tag{10}$$

The z value obtained as a result of this process is passed through the sigmoid function, thus creating a probability value between 0 and 1. The graphical representation of the sigmoid function is shown in Fig. 2.

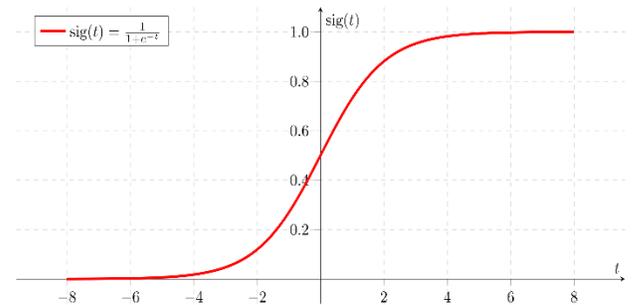

Fig. 2. Graphical representation of sigmoid function

Binary-Cross Entropy is used in the error function [26],[27]. The mathematical expansion of the error function is shown in Equation 11.

$$-[y \ln(\hat{y}) + (1 - y) \ln(1 - \hat{y})] \tag{11}$$

In order to minimize the error function, partial derivatives of the error function are taken with respect to the weights. The partial derivative of the error function with respect to the weights is shown in Equation 12.

$$\frac{\partial L}{\partial w} = \frac{\partial L}{\partial \hat{y}} \cdot \frac{\partial \hat{y}}{\partial z} \cdot \frac{\partial z}{\partial w} \quad (12)$$

The result of the partial derivative of the error function with respect to the weights is shown in Equation 13.

$$\frac{\partial L}{\partial w_k} = (\hat{y} - y) \cdot x_k \quad (13)$$

The formula for updating each weight with the obtained gradients via SGD (Stochastic Gradient Descent) is shown in Equation 14 [28], [29],[30].

$$w_k = w_k - \alpha \frac{\partial L}{\partial w_k} \quad (14)$$

Thus, the weights are updated according to the opposite direction of the derivative. However, the Logistic Regression implementation within EmbeddedML uses the Adam optimizer algorithm and Batch Gradient Descent instead of SGD [31]. Thanks to the Adam optimizer algorithm, the weights are not updated suddenly and learning becomes more stable [32]. Taking into account past gradients by using momentum instead of direct gradient changes of the weights increases the accuracy of the model [33],[34],[35]. While the finding of the moments is shown in Equation 15, the updating of the weights depending on the moments is shown in Equation 16.

$$\vartheta_t = \beta \vartheta_{(t-1)} + (1 - \beta)\nabla_\theta \quad (15)$$
$$w_t = w_{(t-1)} - \alpha \vartheta_t \quad (16)$$

Instead of calculating the error margin of each piece of data in the data set one by one, the cumulative error margin of each piece of data is calculated with Batch Gradient Descent, thus shortening the training time [36],[37]. The mathematical representation of the Batch Gradient Descent algorithm is shown in Equation 17.

$$\theta = \theta - \alpha \nabla J(\theta) \quad (17)$$

With all these mathematical changes made in Logistic Regression, model training time has decreased by 2 times compared to scikit-learn on large datasets.

### 2. KNN

Technically, there is no training in the KNN algorithm because it does not attempt to find any coefficients or weights [38]. Instead, we calculate the Euclidean distance between the data piece whose class we want to predict and each data piece in our train dataset. The class of the data piece we want to predict is estimated based on the number of classes in the data pieces with the least Euclidean distance according to the number of neighbors we choose [39]. The formula that allows us to measure the distance between data pieces in KNN is shown in Equation 18.

$$d = \sqrt{\sum_1^k (x - x_k)^2} \quad (18)$$

### 3. SVM

An attempt is made to determine the most suitable line that separates the data belonging to two different classes in a way that maximizes the distance between them. To maximize this difference, the distance between the support vectors of two different classes is calculated [40],[41]. The formula for the nearest perpendicular distance to the support vectors is shown in Equation 19 and the formula for maximizing this difference is shown in Equation 20.

$$d = \frac{2}{||w||} \quad (19)$$

$$max(\frac{2}{||w||}) = min(\frac{1}{2}||w||^2) \quad (20)$$

The representation of support vectors on the graph in SVM and the measurement of the distance between them are presented in Fig. 3.

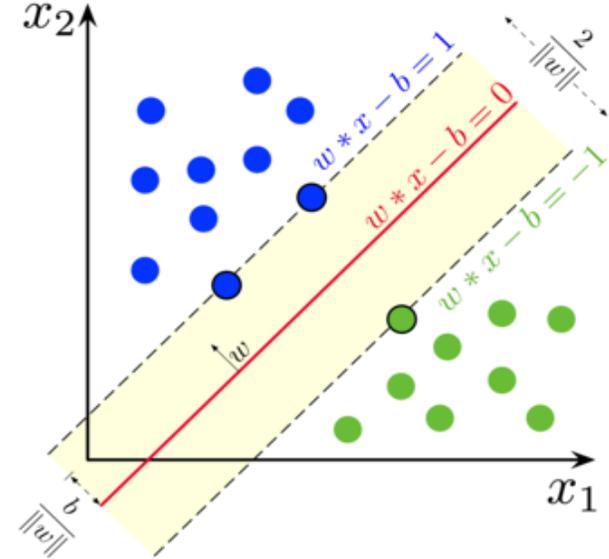

Fig. 3. Graphical representation of the SVM model

In SVM, when calculating the error function, if the predicted class is correct, only the weights are regularized, thus preventing the weights from growing too large and preventing overfitting [42],[43], [44]. If the predicted class is wrong, the weights are updated by both regularizing the weights and adding the error term [45],[46]. The error function used in SVM is shown in Equation 21.

$$\frac{1}{2}\lambda ||w||^2 + \sum_1^k max(0, 1 - y_k(wx_k + b)) \quad (21)$$

If the classification process is estimated, each weight is updated by taking the partial derivative of the error function according to the weights. The method of regularizing the weights and the partial derivatives of the error term is shown in Equations 22 and 23.

$$\frac{1}{2}\lambda ||w||^2 \xrightarrow{partial\ derivative} \lambda w_k \quad (22)$$

$$\sum_1^k max(0, 1 - y_k(wx_k + b)) \xrightarrow{partial\ derivative} -y_k w_k \quad (23)$$

One of the most significant differences between the SVM algorithm in EmbeddedML and scikit-learn is that the weights are updated with momentum, not instantaneous gradients. Updating each weight with momentum prevents sudden changes in weight updates, making learning more stable. Another distinct feature that distinguishes the SVM in EmbeddedML from the SVM algorithm in the scikit-learn library is that the training process is automatically terminated if the number of correct classifications in each epoch exceeds a certain threshold. This prevents unnecessarily long training periods and completes more quickly. Thus, in large datasets, the required accuracy is achieved in a single epoch instead of training in dozens of epochs.

### 4. Naive Bayes

Each data point in the dataset is clustered according to its class. Then, the arithmetic mean and standard deviation values are calculated for each feature of each class. The formula used for standard deviation is shown in Equation 24.

$$\sqrt{\frac{1}{n-1}\sum_1^n (x_n - \bar{x})^2} \quad (24)$$

Thus, each class is defined with the statistical parameters of each feature, completing the training process [47]. For a piece of data to be predicted, each feature is evaluated using a Gaussian Probability Density Function. Gaussian Probability Density Function is shown in Equation 25 [48],[49].

$$P(x_i \mid y) = \frac{1}{\sqrt{2\pi\sigma_y^2}} \exp\left(-\frac{(x_i - \mu_y)^2}{2\sigma_y^2}\right) \quad (25)$$

This process yields a probability value for each feature. Assuming the features are independent, these probability values are multiplied [50]. The formula for the product of independent events used in Naive Bayes is Equation 26.

$$P(x_1) \times P(x_2) \times \cdots \times P(x_n) = \prod_{i=1}^{n} P(x_i) \quad (26)$$

This is the fundamental assumption of the Naive Bayes classifier. These combined probability values for each class are compared, and the class with the highest probability is determined as the predicted label for the relevant data. Additionally, performing such mathematical operations with NumPy has caused a speed increase compared to scikit-learn.

### C. Clustering Algorithm

The K-Means algorithm is used to cluster data consisting of only independent data segments. First, the number of data points to be segmented is determined. Centroids are randomly selected from the dataset based on the determined number of data points. The distance between these selected centroids and each data point in the data set is calculated using the Euclidean distance [51]. Thus, the data points with the shortest distance to each selected centroid are found [52], [53]. This determines clusters. At each iteration, the average of the clusters is calculated, and these averaged values become the new centroids. The distance between the newly identified centroids and the data points in the dataset is again calculated using the Euclidean distance [54]. This way, as the number of iterations increases, the data groups are better separated and grouped. An example visual of the clustering process is illustrated in Fig. 4.

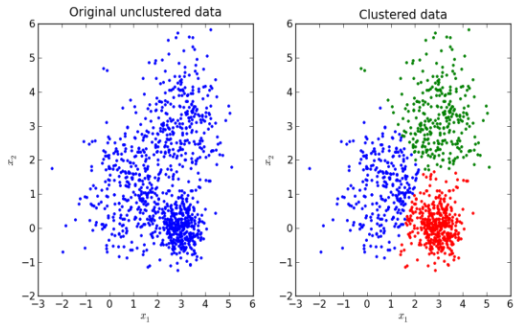

Fig. 4. A graph of the clustering process

### D. Principal Component Analysis (PCA)

Training large and complex datasets can take considerable time. Therefore, dimensionality reduction may be necessary when working with such datasets. To achieve dimensionality reduction, we need information about the distribution of the dataset rather than its location [55]. To obtain this information, we need to eliminate the effect of location on the dataset [56],[57]. For this purpose, we apply centralization to the matrix dataset. Centralization is the process of subtracting the mean of the dataset, setting the mean of each feature (column) to zero, thus moving the dataset to the origin. Centralization is accomplished by calculating the mean of each column and subtracting this mean from each element. The mathematical representation of this process is presented in Equation 27.

$$X_c = X - \bar{X} \quad (27)$$

The covariance matrix of the matrix consisting of centralized and independent data is found [58]. In this way, the interaction of each feature with each other is calculated [59]. The mathematical representation of the covariance matrix is shown in Equation 28.

$$Cov(x) = \frac{1}{n-1}(X_c^T X_c) \quad (28)$$

The eigenvalues and eigenvectors of the resulting covariance matrix are found [60]. Each eigenvector represents the direction of variance propagation in the dataset. Eigenvectors are sorted from largest to smallest based on their eigenvalues [61]. For dimensionality reduction, the eigenvectors with the largest eigenvalues corresponding to the desired dimension are selected and normalized to become unit vectors. Finally, by multiplying these normalized eigenvectors with the centralized data matrix, the dimensionally reduced data, ready for model training, is obtained. To find the preserved variance during the dimension reduction process, the sum of the eigenvalues of each selected eigenvector is divided by the sum of all eigenvalues [62]. The formula for finding the preserved variance is shown in Equation 29.

$$\frac{\lambda_1 + \lambda_2 + \lambda_3 + \cdots + \lambda_j}{\lambda_1 + \lambda_2 + \lambda_3 + \lambda_4 + \lambda_5 + \cdots + \lambda_n} \quad (29)$$

By taking the ratio of the cumulative sum of eigenvalues to the sum of all eigenvalues, dimension reduction can be automatically performed according to the amount of variance that is targeted to be preserved. In the EmbeddedML library, the matrix multiplication, covariance matrix calculation, and eigenvalue and eigenvector finding operations within the PCA (Principal Component Analysis) algorithm are implemented using the NumPy library. This allows for fast and efficient dimensionality reduction.

### E. Preprocessing

Before training the data with machine learning algorithms, some pre-processing techniques are applied, thus increasing the accuracy of the model. The EmbeddedML library includes three basic preprocessing functions: Min-Max Scaler, Standard Scaler, and Train-Val Split. These functions simplify the process of appropriately scaling data before training machine learning models and splitting the dataset into training and validation.

#### 1. Min-Max Scaler

The maximum and minimum values of each column of independent data in matrix format are found. According to these values, the data for each column in the matrix is normalized. The form of the mix-max scaler used in EmbeddedML is shown in Equation 30.

$$X_{\text{normalized}} = \frac{X - X_{\min}}{X_{\max} - X_{\min}} \quad (30)$$

#### 2. Standard Scaler

The arithmetic mean and standard deviation of the data in each column of the matrix are calculated. The arithmetic mean is then subtracted from each column and the result is divided by the standard deviation to normalize the data. The formula for standard deviation used in EmbeddedML is shown in Equation 31.

$$X_{\text{normalized}} = \frac{X - \mu}{\sigma} \quad (31)$$

#### 3. Train-Val Split

To divide the data into train and validation, the data is randomly divided according to the ratio specified by the user, thus determining both train and validation data.

## F. Metrics

Error metrics are used to measure the correct prediction and classification ability of the models as a result of training with machine learning algorithms [63]. In the EmbeddedML library, R² Score, MSE (Mean Squared Error), MAE (Mean Absolute Error), and RMSE (Root Mean Squared Error) metrics are used to evaluate regression models [64]. The R² Score indicates how well the model explains the change in the target variable (dependent variable). It takes a value between 0 and 1. A value close to 1 means the model explains the data very well. MSE (Mean Squared Error) is the average of the squared differences between the actual value and the predicted value. It penalizes large errors more because the differences are calculated by taking the squares. MAE (Mean Absolute Error) is the average of the absolute values of the prediction errors. It shows how far the predictions deviate on average. RMSE (Root Mean Squared Error) is the square root of MSE. Accuracy, precision, recall, and F1 Score metrics are preferred for evaluating classification models [65]. Accuracy measures the overall accuracy of the model, while precision and recall provide more specific information about the model's performance across individual classes. The F1 Score combines precision and recall into a single value by taking their harmonic average. The mathematical representation of error metrics is shown in Table I.

TABLE I. Mathematical Representation Of Error Metrics

| Error Metrics | Mathematical Representation |
|---|---|
| R² Score | $R^2 = 1 - \frac{\sum_{i=1}^{n}(y_i - \hat{y}_i)^2}{\sum_{i=1}^{n}(y_i - \bar{y})^2}$ (32) |
| Mean Squared Error | $MSE = \frac{1}{n}\sum_{i=1}^{n}(y_i - \hat{y}_i)^2$ (33) |
| Mean Absolute Error | $MAE = \frac{1}{n}\sum_{i=1}^{n}|y_i - \hat{y}_i|$ (34) |
| Root Mean Squared Error | $RMSE = \sqrt{\frac{1}{n}\sum_{i=1}^{n}(y_i - \hat{y}_i)^2}$ (35) |
| Accuracy | $Accuracy = \frac{TP+TN}{TP+TN+FP+FN}$ (36) |
| Precision | $Precision = \frac{TP}{TP+FP}$ (37) |
| Recall | $Recall = \frac{TP}{TP+FN}$ (38) |
| F1 Score | $F_1 = 2 \times \frac{Precision \times Recall}{Precision + Recall}$ (39) |

## III. RESULTS

In the EmbeddedML library, machine learning algorithms are written in NumPy from the beginning and mathematically optimized, resulting in a significant increase in speed. To train machine learning models, eight different datasets of varying sizes and content were collected via Kaggle. Eight datasets used in this study are available at https://github.com/HalilHuseyinCaliskan/embeddedML.

The contents of the datasets used for regression models are presented in Table II and the size of the datasets are shown in Table III.

TABLE II. The Content Of The Datasets Used For Regression

| Dataset Number | Dataset Content |
|---|---|
| Dataset 1 | Residential energy consumption |
| Dataset 2 | Computer hardware prices |
| Dataset 3 | Household energy consumption |
| Dataset 4 | Flight ticket prices |

TABLE III. The Sizes Of The Datasets Used For Regression

| Dataset Number | Number of Rows | Number of Columns |
|---|---|---|
| Dataset 1 | 1.000 | 7 |
| Dataset 2 | 6.259 | 9 |
| Dataset 3 | 90.000 | 4 |
| Dataset 4 | 300.153 | 20 |

Table IV provides information about the content of the datasets used to test the classification models, while Table V shows the size of these datasets.

TABLE IV. The Content Of The Datasets Used For Classification

| Dataset Number | Dataset Content |
|---|---|
| Dataset 5 | Diabetes |
| Dataset 6 | Credit Card |
| Dataset 7 | Heart attack |
| Dataset 8 | Thyroid Cancer |

Table V. The sizes of the datasets used for classification models

| Dataset Number | Number of Rows | Number of Columns |
|---|---|---|
| Dataset 5 | 768 | 8 |
| Dataset 6 | 5.000 | 13 |
| Dataset 7 | 100.000 | 15 |
| Dataset 8 | 212.691 | 15 |

Table VI illustrates the data volume complexity of the datasets used for regression. To calculate data volume complexity, the number of rows and the number of columns of each dataset are multiplied.

TABLE VI. The Complexity Of The Datasets Used For Regression And Classification Models

| Dataset Number | Data Volume Complexity |
|---|---|
| Dataset 1 | 7.000 |
| Dataset 2 | 56.331 |
| Dataset 3 | 360.000 |
| Dataset 4 | 6.003.060 |
| Dataset 5 | 6.144 |
| Dataset 6 | 65.000 |
| Dataset 7 | 1.500.000 |
| Dataset 8 | 3.190.365 |

Table VII shows the training times and error metrics measurement results of the Multiple Linear Regression algorithms in the EmbeddedML and scikit-learn libraries on different datasets, and Fig. 5 compares the relationship between dataset volume complexity and training time of Multiple Linear Regression models in EmbeddedML and scikit-learn. According to these test results conducted on a 12th-generation Intel i7 processor, the Multiple Linear Regression model in EmbeddedML reduced the training time by an average of 4 times compared to its counterpart in scikit-learn, with almost no loss in accuracy.

TABLE VII. The Comparison Of The Multiple Linear Regression In EmbeddedML And Scikit-Learn In Terms Of Evaluation Metrics

| Dataset | Evaluation Metrics | EmbeddedML | Scikit-Learn |
|---|---|---|---|
| Dataset 1 | Training Time (ms) | 0.11 ms | 1.04 ms |
| | $R^2$ | 0.9999 | 0.9999 |
| | MSE | 0.0002 | 0.0002 |
| | MAE | 0.0120 | 0.0120 |
| | RMSE | 0.014 | 0.014 |
| Dataset 2 | Training Time (ms) | 0.27 ms | 1.48 ms |
| | $R^2$ | 0.778 | 0.778 |
| | MSE | 78480 | 78480 |
| | MAE | 206.24 | 206.24 |
| | RMSE | 280.14 | 280.14 |
| Dataset 3 | Training Time (ms) | 2.67 ms | 7.36 ms |
| | $R^2$ | 0.9793 | 0.9793 |
| | MSE | 0.6258 | 0.6258 |
| | MAE | 0.6277 | 0.6277 |
| | RMSE | 0.7911 | 0.7911 |
| Dataset 4 | Training Time (ms) | 35 ms | 98 ms |
| | $R^2$ | 0.9038 | 0.9038 |
| | MSE | 49755 | 49755 |
| | MAE | 4693.5 | 4693.5 |
| | RMSE | 7053 | 7053 |

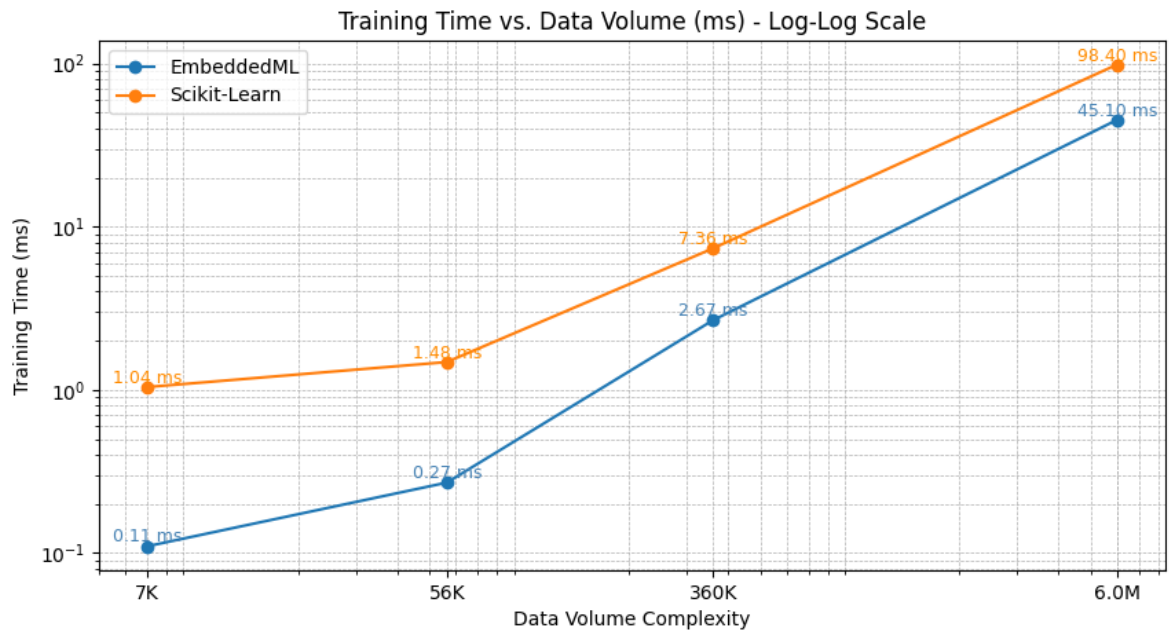

Fig.5 . The Comparison Of The Multiple Linear Regression In EmbeddedML And Scikit-Learn In Terms Of Training Time And Data Volume Complexity

TABLE VIII. The Comparison Of The SVM In Embeddedml And Scikit-Learn In Terms Of Training Time And Error Metrics

| Dataset Type | Evaluation Metrics | EmbeddedML | Scikit-Learn |
|---|---|---|---|
| Dataset 5 | Training Time (ms) | 2.81 ms | 6.40 ms |
| | Accuracy | 74.03 | 73.3 |
| | Precision | 0.55 | 0.70 |
| | Recall | 0.71 | 0.50 |
| | F1 | 0.62 | 0.58 |
| Dataset 6 | Training Time (ms) | 12.5 ms | 166 ms |
| | Accuracy | 72.80 | 74.40 |
| | Precision | 0.80 | 0.79 |
| | Recall | 0.13 | 0.17 |
| | F1 | 0.23 | 0.29 |
| Dataset 7 | Training Time (ms) | 104 ms | 22734 ms |
| | Accuracy | 95.79 | 96.17 |
| | Precision | 0.84 | 0.92 |
| | Recall | 0.62 | 0.60 |
| | F1 | 0.71 | 0.72 |
| Dataset 8 | Training Time (ms) | 345 ms | 278496 ms |
| | Accuracy | 82.78 | 82.78 |
| | Precision | 0.70 | 0.70 |
| | Recall | 0.45 | 0.45 |
| | F1 | 0.55 | 0.55 |

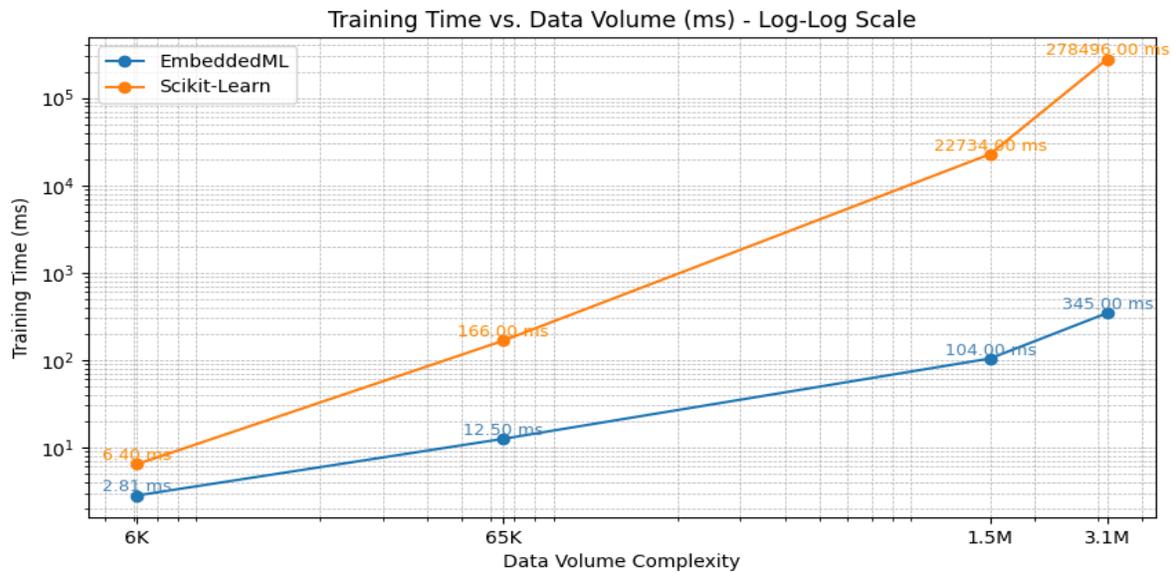

Fig.6 . The Comparison Of The SVM In EmbeddedML And Scikit-Learn In Terms Of Training Time And Data Volume Complexity

Table VIII presents a comparison of the SVM algorithm implemented in EmbeddedML and Scikit-learn in terms of training time and error metrics and, Fig. 6 compares the relationship between dataset volume complexity and training time of SVM algorithms in EmbeddedML and scikit-learn. According to these test results conducted on a 12th-generation Intel i7 processor, the SVM model in EmbeddedML is on average 2 times faster than its scikit-learn counterpart on small datasets and up to approximately 800 times faster on large datasets. The main reasons for this speed increase are that the SVM algorithm is implemented using NumPy and the training process is terminated early at any time when the number of correctly classified examples exceeds a predefined threshold. Despite the speed improvement, the main reason for the negligible loss in accuracy is that the weights are updated using momentum rather than raw gradients in the optimizer algorithm. In this way, the SVM model in EmbeddedML updates the weights by taking into account previous gradi

TABLE IX. The Comparison Of The Logistic Regression In EmbeddedML And Scikit-Learn In Terms Of Training Time And Error Metrics

| Dataset Type | Evaluation Metrics | EmbeddedML | Scikit-Learn |
|---|---|---|---|
| Dataset 5 | Training Time (ms) | 0.28 ms | 0.93 ms |
| | Accuracy | 71.43 | 74.67 |
| | Precision | 0.62 | 0.73 |
| | Recall | 0.60 | 0.51 |
| | F1 | 0.61 | 0.60 |
| Dataset 6 | Training Time (ms) | 1.87 ms | 6.58 ms |
| | Accuracy | 74.7 | 74.5 |
| | Precision | 0.81 | 0.80 |
| | Recall | 0.18 | 0.17 |
| | F1 | 0.299 | 0.293 |
| Dataset 7 | Training Time (ms) | 9.79 ms | 61.1 ms |
| | Accuracy | 94.20 | 96.0 |
| | Precision | 0.63 | 0.86 |
| | Recall | 0.76 | 0.63 |
| | F1 | 0.69 | 0.73 |
| Dataset 8 | Training Time (ms) | 21.7 ms | 65.0 ms |
| | Accuracy | 82.79 | 82.78 |
| | Precision | 0.70 | 0.70 |
| | Recall | 0.45 | 0.45 |
| | F1 | 0.55 | 0.55 |

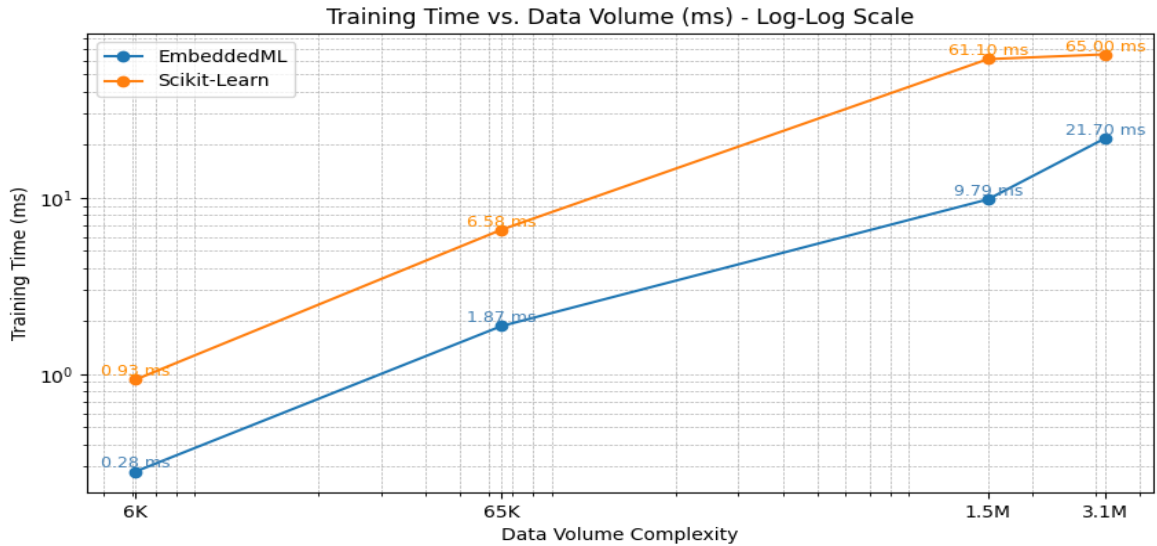

Fig.7 . The comparison of the Logistic Regression in EmbeddedML and scikit-learn in terms of training time and data volume complexity

Table IX presents a comparison of the Logistic Regression algorithm implemented in EmbeddedML and Scikit-learn in terms of training time and error metrics and, Fig. 7 compares the relationship between dataset volume complexity and training time of Logistic Regression algorithms in EmbeddedML and scikit-learn. According to these test results conducted on a 12th-generation Intel i7 processor, the Logistic Regression model in EmbeddedML is on average 4 times faster than its scikit-learn counterpart.

The main reasons for this speed increase are that the Logistic Regression algorithm is implemented using NumPy and instead of Stochastic Gradient Descent (SGD), Batch Gradient Descent (BGD) was employed for optimization. Thus, instead of processing and updating the weights for each individual data sample in every epoch, the entire dataset is processed collectively in batches during each epoch.

## IV. Discussion

In this study, machine learning algorithms developed using the EmbeddedML and Scikit-learn libraries were compared using different error metrics on various datasets. The results show that the EmbeddedML library's rewriting of the models using NumPy support and the mathematical optimizations implemented in the algorithms' internal structures resulted in a significant reduction in training times without any significant loss in model accuracy. It was observed that the training time difference between EmbeddedML and Scikit-learn widened further, particularly as the size and complexity of the datasets increased. This suggests that EmbeddedML may be a more suitable option for real-time or resource-constrained applications in embedded systems.

## V. Conclusion

The EmbeddedML library allows for fast and efficient execution of regression, classification, clustering, and PCA operations on large and complex datasets. This enables rapid training of complex datasets, even on resource-limited embedded systems such as the Raspberry Pi, NVIDIA Jetson Orin, and Orange Pi. In future work, the EmbeddedML library will be expanded beyond statistics-based machine learning algorithms to support more complex neural network-based structures such as CNN, RNN, and LSTM.

## References

[1] James, G., Witten, D., Hastie, T., Tibshirani, R., & Taylor, J. (2023). Linear regression. In *An introduction to statistical learning: With applications in python* (pp. 69-134). Cham: Springer international publishing.

[2] Habehh, H., & Gohel, S. (2021). Machine learning in healthcare. *Current genomics*, *22*(4), 291-300.

[3] Athey, S. (2018). The impact of machine learning on economics. In *The economics of artificial intelligence: An agenda* (pp. 507-547). University of Chicago Press.

[4] Luan, H., & Tsai, C. C. (2021). A review of using machine learning approaches for precision education. *Educational Technology & Society*, *24*(1), 250-266.

[5] Raschka, S., Liu, Y. H., & Mirjalili, V. (2022). *Machine Learning with PyTorch and Scikit-Learn: Develop machine learning and deep learning models with Python*. Packt Publishing Ltd.

[6] Gupta, P., & Bagchi, A. (2024). Introduction to NumPy. In *Essentials of Python for Artificial Intelligence and Machine Learning* (pp. 127-159). Cham: Springer Nature Switzerland.

[7] Molin, S. (2021). *Hands-On Data Analysis with Pandas: A Python data science handbook for data collection, wrangling, analysis, and visualization*. Packt Publishing Ltd.

[8] Hagedorn, S., Kläbe, S., & Sattler, K. U. (2021, January). Putting pandas in a box. In *Conference on Innovative Data Systems Research (CIDR);(Online)* (Vol. 15).

[9] Sial, A. H., Rashdi, S. Y. S., & Khan, A. H. (2021). Comparative analysis of data visualization libraries Matplotlib and Seaborn in Python. *International Journal*, *10*(1), 277-281.

[10] Sarker, I. H. (2021). Machine learning: Algorithms, real-world applications and research directions. *SN computer science*, *2*(3), 160.

[11] Prasad, M., & Srikanth, T. (2024). Clustering Accuracy Improvement Using Modified Min-Max Normalization Technique.

[12] Thakker, Z. L., & Buch, S. H. (2024). Effect of Feature Scaling Pre-processing Techniques on Machine Learning Algorithms to Predict Particulate Matter Concentration for Gandhinagar, Gujarat, India. *Int. J. Sci. Res. Sci. Technol*, *11*(1), 410-419.

[13] Liang, J. (2022). Confusion matrix: Machine learning. *POGIL Activity Clearinghouse*, *3*(4).

[14] Valladares, S., Toscano, M., Tufiño, R., Morillo, P., & Vallejo-Huanga, D. (2021, January). Performance evaluation of the Nvidia Jetson Nano through a real-time machine learning application. In *International Conference on Intelligent Human Systems Integration* (pp. 343-349). Cham: Springer International Publishing.

[15] Kohavi, R., John, G., Long, R., Manley, D., & Pfleger, K. (1994, November). MLC++: A machine learning library in C++. In *Proceedings Sixth International Conference on Tools with Artificial Intelligence. TAI 94* (pp. 740-743). IEEE.

[16] Montgomery, D. C., Peck, E. A., & Vining, G. G. (2021). *Introduction to linear regression analysis*. John Wiley & Sons.

[17] Baždarić, K., Šverko, D., Salarić, I., Martinović, A., & Lucijanić, M. (2021). The ABC of linear regression analysis: What every author and editor should know. *European science editing*, *47*.

[18] Maulud, D., & Abdulazeez, A. M. (2020). A review on linear regression comprehensive in machine learning. *Journal of applied science and technology trends*, *1*(2), 140-147.

[19] Montgomery, D. C., Peck, E. A., & Vining, G. G. (2021). *Introduction to linear regression analysis*. John Wiley & Sons.

[20] Alita, D., Putra, A. D., & Darwis, D. (2021). Analysis of classic assumption test and multiple linear regression coefficient test for employee structural office recommendation. *IJCCS (Indonesian Journal of Computing and Cybernetics Systems)*, *15*(3), 295-306.

[21] Knights, V., & Prchkovska, M. (2024). From equations to predictions: understanding the mathematics and machine learning of multiple linear regression. *J. Math. Comput. Appl*, *3*(2), 1-8.

[22] Das Hait, M., Das, P., Akram, W., & Chatterjee, S. (2025). A Comparative Analysis of Linear Regression Techniques: Evaluating Predictive Accuracy and Model Effectiveness. *International Journal of Innovative Science and Research Technology*, *10*(7), 127-139.

[23] Ostertagová, E. (2012). Modelling using polynomial regression. *Procedia engineering*, *48*, 500-506.

[24] Patil, S., & Patil, S. (2021). Linear with polynomial regression: Overview. *International Journal of Applied Research*, *7*(8), 273-275.

[25] Das, A. (2021). Logistic regression. In *Encyclopedia of quality of life and well-being research* (pp. 1-2). Cham: Springer International Publishing.

[26] Boateng, E. Y., & Abaye, D. A. (2019). A review of the logistic regression model with emphasis on medical research. *Journal of data analysis and information processing*, *7*(04), 190.

[27] Mao, A., Mohri, M., & Zhong, Y. (2023, July). Cross-entropy loss functions: Theoretical analysis and applications. In *International conference on Machine learning* (pp. 23803-23828). pmlr.

[28] Manogaran, G., & Lopez, D. (2018). Health data analytics using scalable logistic regression with stochastic gradient descent. *International Journal of Advanced Intelligence Paradigms*, *10*(1-2), 118-132.

[29] Zou, X., Hu, Y., Tian, Z., & Shen, K. (2019, October). Logistic regression model optimization and case analysis. In *2019 IEEE 7th international conference on computer science and network technology (ICCSNT)* (pp. 135-139). IEEE.

[30] Du, J. (2019, May). The frontier of SGD and its variants in machine learning. In *Journal of Physics: Conference Series* (Vol. 1229, No. 1, p. 012046). IOP Publishing.

[31] Tato, A., & Nkambou, R. (2018). Improving adam optimizer.

[32] Ye, M., Yan, X., Chen, N., & Liu, Y. (2024). A robust multi-scale learning network with quasi-hyperbolic momentum-based Adam optimizer for bearing intelligent fault diagnosis under sample imbalance scenarios and strong noise environment. *Structural Health Monitoring*, *23*(3), 1664-1686.

[33] Liu, W., Chen, L., Chen, Y., & Zhang, W. (2020). Accelerating federated learning via momentum gradient descent. *IEEE Transactions on Parallel and Distributed Systems*, *31*(8), 1754-1766.

[34] Huang, F., Gao, S., Pei, J., & Huang, H. (2020, November). Momentum-based policy gradient methods. In *International conference on machine learning* (pp. 4422-4433). PMLR.

[35] Ramezani-Kebrya, A., Antonakopoulos, K., Cevher, V., Khisti, A., & Liang, B. (2024). On the generalization of stochastic gradient descent with momentum. *Journal of Machine Learning Research*, *25*(22), 1-56.

[36] Ruder, S. (2016). An overview of gradient descent optimization algorithms. *arXiv preprint arXiv:1609.04747*.

[37] Wang, X., Yan, L., & Zhang, Q. (2021, September). Research on the

[37] application of gradient descent algorithm in machine learning. In *2021 international conference on computer network, electronic and automation (ICCNEA)* (pp. 11-15). IEEE.

[38] Zhang, S., Li, X., Zong, M., Zhu, X., & Cheng, D. (2017). Learning k for knn classification. *ACM Transactions on Intelligent Systems and Technology (TIST)*, *8*(3), 1-19.

[39] Zhang, S. (2021). Challenges in KNN classification. *IEEE Transactions on Knowledge and Data Engineering*, *34*(10), 4663-4675.

[40] Chauhan, V. K., Dahiya, K., & Sharma, A. (2019). Problem formulations and solvers in linear SVM: a review. *Artificial Intelligence Review*, *52*(2), 803-855.

[41] Pisner, D. A., & Schnyer, D. M. (2020). Support vector machine. In *Machine learning* (pp. 101-121). Academic Press.

[42] Ghosh, S., Dasgupta, A., & Swetapadma, A. (2019, February). A study on support vector machine based linear and non-linear pattern classification. In *2019 International conference on intelligent sustainable systems (ICISS)* (pp. 24-28). IEEE.

[43] Loshchilov, I., & Hutter, F. (2017). Decoupled weight decay regularization. *arXiv preprint arXiv:1711.05101*.

[44] Van Laarhoven, T. (2017). L2 regularization versus batch and weight normalization. *arXiv preprint arXiv:1706.05350*.

[45] Xu, G., Cao, Z., Hu, B. G., & Principe, J. C. (2017). Robust support vector machines based on the rescaled hinge loss function. *Pattern Recognition*, *63*, 139-148.

[46] Kavalerov, I., Czaja, W., & Chellappa, R. (2021). A multi-class hinge loss for conditional gans. In *Proceedings of the IEEE/CVF winter conference on applications of computer vision* (pp. 1290-1299).

[47] Yang, F. J. (2018, December). An implementation of naive bayes classifier. In *2018 International conference on computational science and computational intelligence (CSCI)* (pp. 301-306). IEEE.

[48] Reddy, E. M. K., Gurrala, A., Hasitha, V. B., & Kumar, K. V. R. (2022). Introduction to Naive Bayes and a review on its subtypes with applications. *Bayesian reasoning and gaussian processes for machine learning applications*, 1-14.

[49] Wickramasinghe, I., & Kalutarage, H. (2021). Naive Bayes: applications, variations and vulnerabilities: a review of literature with code snippets for implementation. *Soft Computing*, *25*(3), 2277-2293.

[50] Kamel, H., Abdulah, D., & Al-Tuwaijari, J. M. (2019, June). Cancer classification using gaussian naive bayes algorithm. In *2019 international engineering conference (IEC)* (pp. 165-170). IEEE.

[51] Jin, X., & Han, J. (2017). K-means clustering. In *Encyclopedia of machine learning and data mining* (pp. 695-697). Springer, Boston, MA.

[52] Ahmed, M., Seraj, R., & Islam, S. M. S. (2020). The k-means algorithm: A comprehensive survey and performance evaluation. *Electronics*, *9*(8), 1295.

[53] Chong, B. (2021). K-means clustering algorithm: a brief review. *Academic Journal of Computing & Information Science*, *4*(5), 37-40.

[54] Sinaga, K. P., & Yang, M. S. (2020). Unsupervised K-means clustering algorithm. *IEEE access*, *8*, 80716-80727.

[55] Kabari, L. G., & Nwamae, B. B. (2019). Principal component analysis (pca)-an effective tool in machine learning. *Int. J. Advanced Research in Computer Science and Software Engineering*, *9*(5), 56-59.

[56] Sudharsan, M., & Thailambal, G. (2023). Alzheimer's disease prediction using machine learning techniques and principal component analysis (PCA). *Materials Today: Proceedings*, *81*, 182-190.

[57] Swathi, P., & Pothuganti, K. (2020). Overview on principal component analysis algorithm in machine learning. *International Research Journal of Modernization in Engineering Technology and Science*, *2*(10), 241-246.

[58] Johnstone, I. M., & Paul, D. (2018). PCA in high dimensions: An orientation. *Proceedings of the IEEE*, *106*(8), 1277-1292.

[59] Bloemendal, A., Knowles, A., Yau, H. T., & Yin, J. (2016). On the principal components of sample covariance matrices. *Probability theory and related fields*, *164*(1), 459-552.

[60] Ghojogh, B., Karray, F., & Crowley, M. (2019). Eigenvalue and generalized eigenvalue problems: Tutorial. *arXiv preprint arXiv:1903.11240*.

[61] Montano, V., & Jombart, T. (2017). An Eigenvalue test for spatial principal component analysis. *BMC bioinformatics*, *18*(1), 562.

[62] Jolliffe, I. T., & Cadima, J. (2016). Principal component analysis: a review and recent developments. *Philosophical transactions of the royal society A: Mathematical, Physical and Engineering Sciences*, *374*(2065), 20150202.

[63] Naidu, G., Zuva, T., & Sibanda, E. M. (2023, April). A review of evaluation metrics in machine learning algorithms. In *Computer science on-line conference* (pp. 15-25). Cham: Springer International Publishing.

[64] Rainio, O., Teuho, J., & Klén, R. (2024). Evaluation metrics and statistical tests for machine learning. *Scientific Reports*, *14*(1), 6086.

[65] Van Thieu, N. (2024). Permetrics: A framework of performance metrics for machine learning models. *Journal of Open Source Software*, *9*(95), 6143.

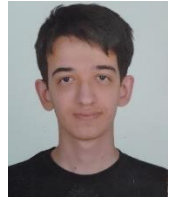

**Halil Hüseyin Çalışkan** is an undergraduate student in the Department of Computer Engineering at Bursa Technical University's Faculty of Engineering and Natural Sciences. His current interests include AI-based solutions for real-world problems and optimizing machine learning algorithms.

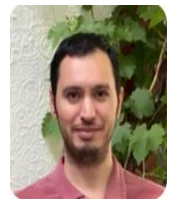

**Talha Koruk** received the B.Sc. and M.Sc. degrees in Computer Engineering from the Middle East Technical University in 2012 and 2018, respectively. He is currently pursuing the Ph.D. degree with the Department of Computer Engineering, Bursa Technical University. His research interests include machine learning, spiking neural networks and anomaly detection. His research interests include machine learning and anomaly prediction areas with practical effects on real-world applications.

.